\documentclass[10pt,twocolumn,letterpaper]{article}

\usepackage[pagenumbers]{cvpr} 

\usepackage{booktabs}       
\usepackage{siunitx}        
\usepackage{multirow}       
\usepackage{amsmath}        
\usepackage{array} 
\usepackage{graphicx} 
\usepackage{amssymb}        

\usepackage[dvipsnames]{xcolor}
\usepackage{float}
\usepackage{makecell}
\usepackage{pifont}
\usepackage{listings}
\usepackage{enumitem}
\usepackage{color, colortbl}
\usepackage{wrapfig}

\usepackage{etoc}
\usepackage{titletoc}
\usepackage{wrapfig}
\usepackage{cuted}
\usepackage{caption, subcaption, overpic, textpos}

\usepackage{comment}
\usepackage{lipsum}
\usepackage{bbding}
\usepackage{diagbox}

\usepackage{algorithm}
\usepackage{listings}
\usepackage{etoolbox}

\definecolor{mblue}{HTML}{6895D2}
\definecolor{morange}{HTML}{E48F45}
\definecolor{mpurple}{HTML}{836096}
\definecolor{mpink}{HTML}{CD6688}

\definecolor{cvprblue}{rgb}{0.21,0.49,0.74}
\usepackage[pagebackref,breaklinks,colorlinks,citecolor=cvprblue]{hyperref}

\def\paperID{3209} 
\def\confName{CVPR}
\def\confYear{2025}

\title{GaussianPretrain: A Simple Unified 3D Gaussian Representation for \\ Visual Pre-training in Autonomous Driving}

\author{
Shaoqing Xu$^{1}$ \hspace{0.35cm} 
Fang Li$^2$ \hspace{0.25cm} Shengyin Jiang$^3$ \hspace{0.25cm} Ziying Song$^4$ \hspace{0.25cm} 
Li Liu$^2$ \hspace{0.25cm}  Zhi-xin Yang$^{1\dagger}$ \\
$^1$University of Macau \hspace{0.9cm} $^2$Beijing Institute of
Technology \hspace{0.9cm} \\ $^3$Beijing University of Posts 
and Telecommunications \hspace{0.9cm} $^4$Beijing Jiaotong University \\
}

\begin{document}
\twocolumn[{%
\renewcommand\twocolumn[1][]{#1}%
\maketitle
\vspace{-25pt}
\begin{center}
\centering
\includegraphics[width=1\linewidth]{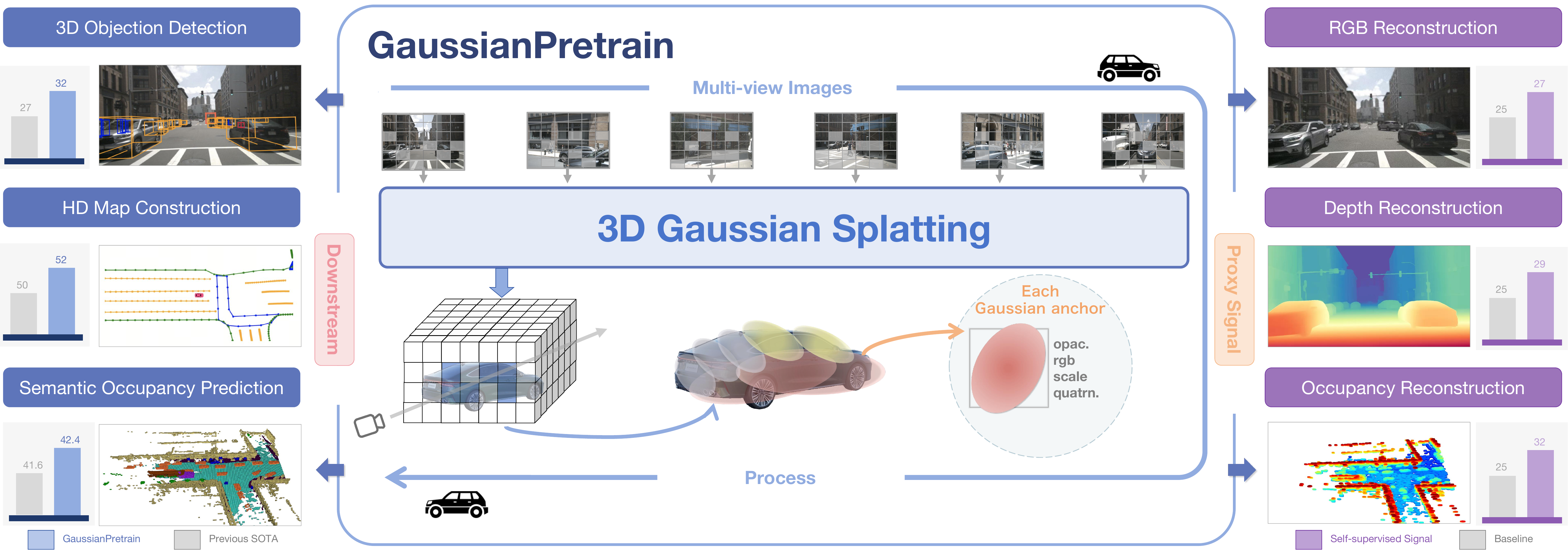}
\label{fig:head_fig}
\captionof{figure}{Illustration of our proposed GaussianPretrain. A simple, innovative, and efficient framework for vision pre-training with 3D Gaussian splitting (3D-GS) representation. Benefits from our effective pre-training diagram, downstream application for 3D perception tasks achieved great improvement, including 3D object detection, HD-map construction, and Occupancy prediction.
}
\end{center}%
}]

\begin{abstract}
Self-supervised learning has made substantial strides in image processing, while visual pre-training for autonomous driving is still in its infancy. Existing methods often focus on learning geometric scene information while neglecting texture or treating both aspects separately, hindering comprehensive scene understanding. In this context, we are excited to introduce \textbf{GaussianPretrain}, a novel pre-training paradigm that achieves a holistic understanding of the scene by uniformly integrating geometric and texture representations. Conceptualizing 3D Gaussian anchors as volumetric LiDAR points, our method learns a deepened understanding of scenes to enhance pre-training performance with detailed spatial structure and texture, achieving that \textbf{40.6\%} faster than NeRF-based method UniPAD with \textbf{70\%} GPU memory only. 
We demonstrate the effectiveness of GaussianPretrain across multiple 3D perception tasks, showing significant performance improvements, such as a \textbf{7.05\%} increase in NDS for 3D object detection, boosts mAP by \textbf{1.9\%} in HD map construction and {0.8\%} improvement on Occupancy prediction. These significant gains highlight GaussianPretrain’s theoretical innovation and strong practical potential, promoting visual pre-training development for autonomous driving.
Source code will be available at {\textcolor{magenta}{\url{https://github.com/Public-BOTs/GaussianPretrain}}}
\end{abstract}
    
\vspace{-1em}
\section{Introduction}
\label{sec:intro}
With the development of autonomous driving technology, vision-centered solutions have gradually attracted widespread attention. Many studies focus on extracting bird's-eye view (BEV) features from multi-view input images to address various downstream applications. While supervised methods dominate current research, their reliance on accurate ground truth labels presents a significant bottleneck due to the high cost and difficulty of acquisition. Conversely, the abundance and accessibility of unlabeled data offer a promising avenue for improving performance. However, effectively harnessing this unlabeled data remains a significant and ongoing challenge in the field.

The core idea of self-supervised pre-training technology is to learn meaningful representations from abundant unlabeled data by leveraging carefully designed proxy tasks. 
There are several approaches have been developed to explore this topic,
methods like UniScene~\cite{min2023uniscene} and  ViDAR~\cite{yang2024visual} focus on predicting 3D occupancy or future LiDAR point clouds, effectively capturing geometric information but neglecting texture. Conversely,
Self-Occ~\cite{huang2024selfocc} and UniPAD~\cite{yang2024unipad} reconstruct 3D surfaces and RGB pixels from image-LiDAR pairs, thus capturing texture but relying solely on depth maps for geometric insights. Although this aids in texture learning, these methods remain limited in extracting comprehensive geometric information.
OccFeat~\cite{sirko2024occfeat}, combines feature distillation with occupancy prediction to effectively capture both texture and geometry, but it introduces the complexity of an additional image foundation model and incurs significant pre-training costs.


Fortunately, 3D Gaussian Splatting (3D-GS), represented as point clouds, offers a powerful representation for scene reconstruction, encoding \textbf{geometric} and \textbf{texture} information through attributes like \textit{position, color, rotation, scaling, and opacity}. Furthermore, compared to NeRF, 3D-GS achieves faster training convergence and requires less memory. These advantages address the key limitations in existing 3D pre-training techniques.

%
Inspired by the success of 3D-GS in effective scene representation and MAE~\cite{he2022masked} in 2D image self-supervised learning, we propose a novel pre-training approach \textbf{GaussianPretrain}, which combines 3D-GS with MAE method for pre-training tasks in 3D visual learning.
Our approach incorporates two key innovations: (i) \textbf{\textit{LiDAR Depth Guidance Mask Generator}}. To enhance the efficiency of our approach,  we only focus on learning the Gaussian information from a limited number of valid masked patches within the multi-view images. These patches are identified by an MAE strategy and further filtered to include only those with LiDAR depth supervision. (ii). \textbf{\textit{Ray-based 3D Gaussian anchor Guidance Strategy}}:
For each LiDAR-projected pixel, a ray-casting operation into 3D space to sample the points within the voxel. We introduce a set of learnable Gaussian anchors of these points to guide the learning of  Gaussian properties from the 3D voxel as volumetric LiDAR points and predict the relevant attributes (e.g., \textit{depth, opacity}). This enables the model to simultaneously understand the geometry and texture information of the scene through 3D Gaussian Splatting.
Finally, we reconstruct the \textit{RGB}, \textit{depth}, and \textit{occupancy} attributes solely within the valid masked patches by decoding the Gaussian parameters.

Compared to NeRF~\cite{mildenhall2021nerf} based methods like UniPAD~\cite{yang2024unipad}, our GaussianPretrain offers lower training time costs and memory consumption, while providing well understanding of the scene's representation. By combining these advantages, our GaussianPretrain enables to perform effective and efficient self-supervised pre-training.

\begin{figure}[t!]
  \centering
  \includegraphics[width=1.0\linewidth]{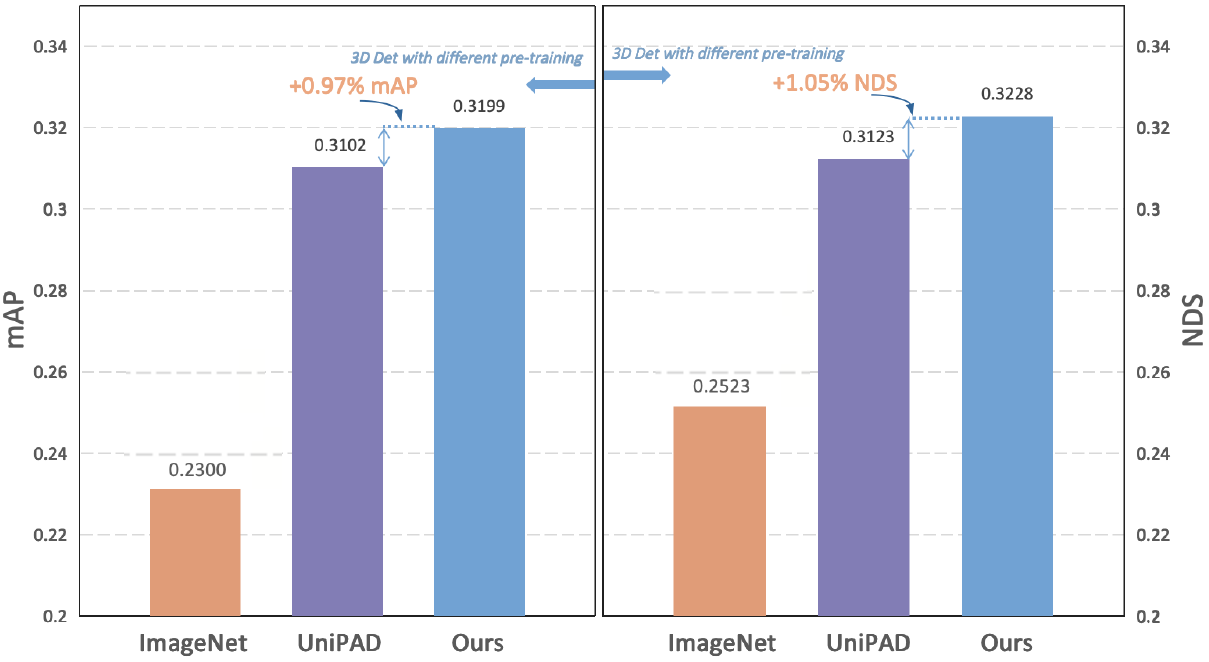}
  \caption{
  Comparison of UVTR~\cite{li2022unifying} model performance on the nuScenes dataset with different pre-training framework: ImageNet, UniPAD~\cite{yang2024unipad}, and our GaussianPretrain.}
  \vspace{-1.2em}
  \label{fig:top}
\end{figure}


To demonstrate the effectiveness and generalizability of GaussianPretrain, we conduct extensive experiments on various downstream vision tasks using the large-scale nuScenes\cite{caesar2020nuscenes} dataset. 
As shown in Figure~\ref{fig:top}, for 3D object detection, our pre-trained model significantly outperforms the ImageNet pre-trained UVTR baseline, achieving the gain in \textbf{8.99\%} mAP and \textbf{7.05\%} NDS. Furthermore, our approach surpasses the previous state-of-the-art method, UniPAD~\cite{yang2024unipad}, by 0.97\% mAP and 1.05\% NDS. 
To further validate the generalizability, we evaluate our method on HD map construction and occupancy prediction tasks. GaussianPretrain surpasses the state-of-the-art MapTR~\cite{liao2022maptr} method for HD map construction by \textbf{1.9\%} in mAP. For occupancy prediction, our approach achieves a {0.8\%}  improvement in mIoU over PanoOCC, setting a new SOTA performance. Further details are provided in \Cref{sec:experiments}.
To summarize, our main contributions are as follows:
\begin{itemize}
    \item We introduce GaussianPretrain, a novel pre-training framework that integrates 3D Gaussian Splatting technology with a unified Gaussian representation. To the best of our knowledge, this is the first work to leverage 3D-GS for pre-training diagram, representing a novel contribution to this field.
    \item We propose a simple yet effective framework by leveraging 3D Gaussian anchors as volumetric LiDAR points, combined with Ray-based guidance and MAE method. Providing an efficient solution for visual pre-training, significantly reduces time consumption and GPU memory costs without sacrificing detail.
    \item The comprehensive experimental results demonstrate the superiority and generalizability of GaussianPretrain, showcasing significant improvements across various 3D perception tasks, including 3D object detection, HD-map construction, and Occupancy prediction. These advancements set new performance standards and underscore GaussianPretrain's transformative potential in visual pre-training tasks for autonomous driving applications.
\end{itemize}

\section{Related Work}
\subsection{Pre-training for Autonomous Driving.}
Pre-training on 2D images has been highly successful, mainly using contrastive learning and masked signal modeling to capture semantic and texture information. However, pre-training for visual autonomous driving demands accurate geometric representation, introducing additional challenges. Several works are currently exploring this area. For instance, UniScene\cite{min2023uniscene} and OccNet\cite{tong2023scene} leverage occupancy prediction for pre-training, while ViDAR~\cite{yang2024visual} predicts future LiDAR data from historical frame images. Although these methods are effective at capturing geometric information, but cannot learn detailed texture information. Conversely, methods like Self-OCC~\cite{huang2024selfocc}, UniPAD~\cite{yang2024unipad}, and MIM4D~\cite{zou2024mim4d} use NeRF~\cite{mildenhall2021nerf} to render RGB images and depth maps, learning texture but with limited geometric information. OccFeat~\cite{sirko2024occfeat} employs knowledge distillation to transfer texture information from an image foundation model during occupancy prediction but incurs high pre-training costs. In contrast, our work introduces 3D Gaussian Splatting for vision pre-training in autonomous driving, efficiently capturing both texture and geometry to address these limitations.

\subsection{3D Gaussian Splatting and NeRF.} 
Neural Radiance Fields (NeRF)~\cite{mildenhall2021nerf} achieve impressive rendering quality by implicitly representing scenes of color and density, parameterized by Multi-Layer Perceptrons (MLPs) combined with volume rendering techniques. Follow-up works \cite{boss2021nerd,park2021nerfies,wang2024perf,xian2021space,irshad2023neo} have successfully extended NeRF to various tasks, these methods still require extensive per-scene optimization, limiting their efficiency due to slow optimization and rendering speeds. In contrast, 3D Gaussian Splatting~\cite{kerbl20233d} explicitly represents scenes with anisotropic Gaussians, enabling real-time rendering via differentiable rasterization. However, it tends to overfit specific scenes due to its reliance on scene-specific optimization. Recent approaches mitigate this by predicting Gaussian parameters in a feed-forward manner, eliminating the need for per-scene optimization. For example, GPS-Gaussian~\cite{zheng2024gps} performs epipolar rectification and disparity estimation from image pairs, relying on stereo images and ground-truth depth maps. Similarly, Spatter Image~\cite{szymanowicz2024splatter} focuses on single-object 3D reconstruction from single views. Both methods are often constrained by inefficiencies, limited to object reconstruction, and dependent on specific input formats such as image pairs or single views. In this paper, we extend 3D Gaussian Splatting into visual pre-training tasks, overcoming limitations related to the number of views and the necessity of depth maps by presetting fixed-position 3D Gaussian anchors in 3D space,  marking a novel application of 3D-GS.

\begin{figure*}[ht]
  \centering
  \includegraphics[width=0.9\linewidth]{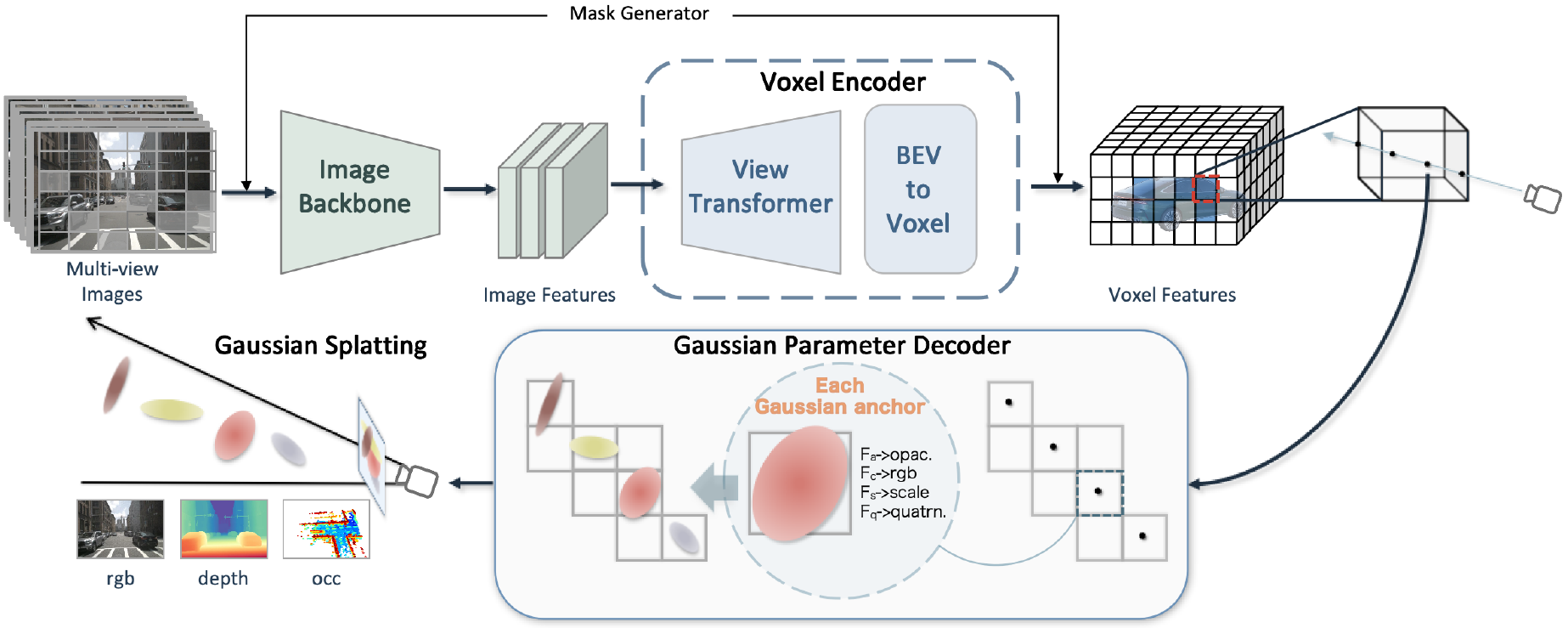}
  \caption{The architecture of proposed GaussianPretrain. Given multi-view images, we first extract valid mask patches using the mask generator with the LiDAR Depth Guidance strategy. Subsequently, a set of learnable 3D Gaussian anchors is generated using ray-based guidance and conceptualized as volumetric LiDAR points. Finally, the reconstruction signals of RGB, Depth, and Occupancy are decoded based on the predicted Gaussian anchor parameters.}
  \label{fig:framework}
  \vspace{-1.0em}
\end{figure*}

\section{Preliminary}
\label{sec:formatting}

\subsection{3D Gaussian Splatting.}
3D Gaussian Splatting~\cite{kerbl20233d} demonstrates strong capabilities in scene representation, editing, and novel view synthesis due to its efficient rasterization design and explicit representation. Generally, scenes are represented by a set of Gaussians which are typically initialized from point clouds derived from reconstruction methods or LiDAR data. Each Gaussian is assigned with learnable attributes of \textit{orientation, color, position, scale}, and \textit{opacity}. During rendering, these 3D Gaussians are projected onto the 2D image plane using differentiable rasterization. For scene representation tasks, the Gaussian attributes are iteratively optimized by supervising the rendered output with ground truth images.

For a Gaussian point in the 3D space, it is defined as
\begin{equation}
G(\mathbf{x}) = e^{-\frac{1}{2}(\mathbf{x} - \boldsymbol{\mu})^\top \boldsymbol{\Sigma}^{-1} (\mathbf{x} - \boldsymbol{\mu})}
\end{equation}
with $\mu$ and $\Sigma$ referring to the Gaussian mean center and the 3D covariance matrix. Projected into 2D, we have the 2D covariance matrix $\Sigma' = J W \Sigma W^T J^T$, where the viewing transformation is represented by 
$W$, and the Jacobian matrix $J$ corresponds to the linear approximation of the transformation. Finally, the pixel color is rendered from N ordered Gaussians with the blending equation
\begin{equation}
\label{reder_rgb}
{C}(p) = \sum_{i=1}^{N} c_i \alpha_i \tau
\end{equation}
where $c_i$ is the Gaussian color represented with spherical harmonics and $\alpha_i$ is the opacity related influence of this Gaussian to the current pixel. $\tau = \prod_{j=1}^{i-1} (1 - \alpha_j)$ is the transmittance.

\section{Method}
The pipeline of our GaussianPretrain, a simple, innovative, and efficient framework for vision pre-training using 3D-GS representation, is illustrated in Figure.\ref{fig:framework}. 
%
Given multi-view images with valid masked patches, 
our goal is to reconstruct the signals, including \textit{RGB}, \textit{Depth}, and \textit{occupancy} by decoding Gaussian parameters $\{ ({\mu}_j, \alpha_j, {\Sigma}_j, {c}_j) \}_{j=1}^{K}$ for each scene, where ${\mu}_j$, $\alpha_j$, ${\Sigma}_j$ and ${c}_j$ are the 3D Gaussian's position, opacity, covariance, and color information, and $K$ denote the max number of Gaussian Anchors. 

%
In this section, we first detail the generation of valid masked patches under MAE method and Depth guidance in ~\Cref{sec:mask_generator}. Then, we introduce the initialization of 3D Gaussian anchors through Ray-based Guidance in ~\Cref{sec:anchor_generator}. Next, we review the process of converting multi-view images into a 3D voxel space in ~\Cref{sec:voxel_encoder}. Subsequently, we present the process of the Gaussian decoder part in \Cref{sec:gaussian_decoder}. Finally, the reconstruction task and GaussianPretrain' loss functions are present in ~\Cref{sec:losses}.

\subsection{\textbf{LiDAR Depth Guidance Mask Generator}}
\label{sec:mask_generator}


Inspired by MAE~\cite{he2022masked}, we apply random patch masking to multi-view images, denote as $M$. Furthermore, sparse convolution is used to replace traditional convolutions in the image backbone, as implemented by SparK~\cite{tian2023designing} which enhances the performance and generalization. For computational efficiency, we only focus on learning Gaussian parameters from a limited set of valid masked patches.

\begin{figure}[t!]
  \centering
  \includegraphics[width=0.8\linewidth]{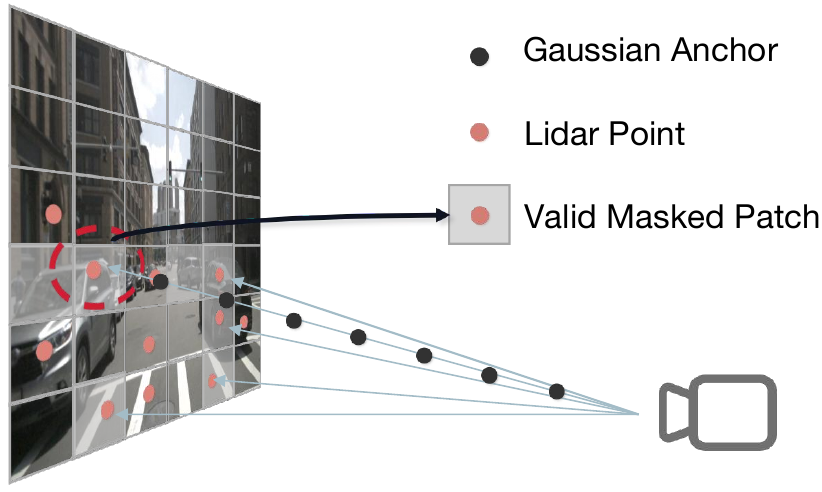}
  \caption{
  \textbf{Process of generating valid mask patches. } 
 }
  \label{fig:motivation}
\end{figure}

Additionally, we double-check the mask region by verifying the presence of LiDAR points within a certain depth range. 
As illustrated in \Cref{fig:motivation}, if a set of points project into the masked patch $M_i$ in the images and their depth falls within the range of [a, b], the mask region will be marked as valid, $M'_i$. The process is outlined as follows:
\begin{equation}
{M'}_{i=1}^n = 
valid,  \text{if } Proj(\text{\textit{{Set}(\textit{pc})}}) \in \{[a, b], M\}
\end{equation} 

where n represents the number of valid masked patches, with n $\leq$ m. 
This strategy ensures that our model concentrates on the foreground, avoiding unnecessary attention to irrelevant background elements like the sky.

\subsection{\textbf{Ray-based Guidance 3D Gaussian Anchor}}
\label{sec:anchor_generator}

To enable the model to simultaneously understand the geometry and texture information of the scene, we introduce a series of learnable Gaussian anchors in 3D space. These anchors guide the learning of Gaussian properties derived from the 3D voxel grid, treated as volumetric LiDAR points.
Consider the LiDAR-projected pixel denoted by $\mathbf{u}=(u_1, u_2, 1)$ which corresponds to a ray $R$ that extends from the camera into 3D space. Along this ray, we sample $D$ ray points $\{p_{j}=\mathbf{u} d_{j}|j=1,...D,\  d_{j}<d_{j+1}\}$, where $d_{j}$ is the corresponding depth along the ray. Each sampled ray point $p$ in the valid mask region $M'$ can be immediately unprojected to 3D space using the projection matrix summary as 3D Gaussian Anchors, $\mathcal{G}_p^{M'}(\cdot)$. 
This strategy not only eliminates the need for full image rendering, significantly reducing memory usage, but also enables simultaneous the RGB, depth, and occupancy reconstruction, a capability that has yet to be achieved by prior methods.
%

\subsection{Voxel Encoder}
\label{sec:voxel_encoder}
{In most perception tasks, view transformer is typically used to generate Bird’s Eye View (BEV) features, which is then utilized in subsequent downstream tasks. 
Notably, our pre-training method is compatible with any type of view transformer.
%
In our baseline model, UVTR~\cite{li2022unifying}, we employ the lift-splat-shoot (LSS)\cite{philion2020lift} and extend the channel dimension to incorporate a height dimension, producing 3D voxel features $V\in R^{C \times Z\times H\times W}$, where $C$, $H$, $W$, and $Z$ represent the channel number, dimensions along the $x$, $y$, and $z$ axes, respectively.
}
Additionally, for each LiDAR-projected pixel, we perform a ray-casting operation to extract $N_t$ sampled target voxel where exists Gaussian Anchors $V_t$ from 3D voxel grid $V$. 

\subsection{\textbf{Gaussian Parameter Decoder}}
\label{sec:gaussian_decoder}

As shown in \Cref{fig:framework}, by conceptualizing $\mathcal{G}_p^{M'} $  as 3D Gaussian anchors, 
this unified representation enables the efficient capture of high-quality, fine-grained details, providing a more comprehensive understanding of the scene.

%
Specially, each 3D  Gaussian anchor is characterized by attributes 
$\mathcal{G}=\{ x \in \mathbb{R}^{3} ,c\in \mathbb{R}^{3} ,r\in \mathbb{R}^{4},s\in \mathbb{R}^{3} ,\alpha \in \mathbb{R}^{1} \}$
and the proposed Gaussian maps $G$ are defined as:
\begin{equation}
G\left( x\right)  =\{ \mathcal{M}_{c} (x),\mathcal{M}_{r} (x),\mathcal{M}_{s} (x),\mathcal{M}_{\alpha } (x)\} 
\end{equation}
where $x$ is the position of a Gaussian anchor in 3D space, $\mathcal{M}_{c}$, $\mathcal{M}_{r}$, $\mathcal{M}_{s}$, $\mathcal{M}_{\alpha }$ represents Gaussian parameters maps of color, rotation, scaling and opacity, respectively.


%
Due to the overlapping areas in the multi-view images, the pixel-by-pixel prediction of Gaussian parameters 
may lead to ambiguity due to overlapping splats. In contrast, we argue that predicting Gaussian parameters
in a feed-forward manner directly from 3D voxel features is a better choice. Given voxel features $V$ and center coordinate $x$ , we employ trilinear interpolation to sample the corresponding feature $f(x)$ as follows:
\begin{equation}
f(x)=TriInter(V,x)
\end{equation}
The Gaussian parameter maps are generated by prediction heads,  defined as $h=MLP(\cdot)$, which consist of multiple MLP layers. Each prediction head is specifically designed to regress a particular parameter based on the sampled feature $f(x)$. For the  parameter of color and opacity, we employed the sigmoid function for a range of [0,1] as follows:
\begin{align}
    \mathcal{M}_{c} (x)=Sigmoid(h_{c}(f(x)) \\
    \vspace{1em}
    \mathcal{M}_{\alpha} (x)=Sigmoid(h_{\alpha}(f(x))
\end{align}
where $h_{c}$, $h_{\alpha}$ denote the head of color and opacity. 

Before being used to formulate Gaussian representations, the rotation map should be normalized since it represents a quaternion to ensure unit magnitude while the scaling map needs activations to satisfy their range as follows:
\begin{align}
    \mathcal{M}_{r} (x)=Norm(h_{r}(f(x))\\
    \vspace{1em}
    \mathcal{M}_{s} (x)=Softplus(h_{s}(f(x)) 
\end{align}
where $h_{r}$, $h_{s}$ represents the rotation head and scale head.

\begin{table*}[t]
	\centering
\resizebox{\textwidth}{!}{%
        \begin{tabular}{l|c|c|cc|c|c|c|c|c}
		\toprule
		Methods & Present at & Backbone & NDS$\uparrow$ & mAP$\uparrow$  & \textbf{{mATE$\downarrow$}} & 
         \textbf{{mASE$\downarrow$}} & 
         \textbf{{mAOE$\downarrow$}} & 
         \textbf{{mAVE$\downarrow$}} & 
         \textbf{{mAAE$\downarrow$}} \\
		\midrule
		BEVFormer-S~\cite{li2022bevformer} & ECCV'22 & R101& 44.8 & 37.5 & -& -& -& -& -\\
		UVTR-C~\cite{li2022unifying} & NeurIPS'22  &R101 & 44.1 & 37.2 & 0.735 & 0.269 & 0.397 & 0.761 &0.193\\        
		PETR~\cite{liu2022petr} & ECCV'22 & R101& 44.2 & 37.0 & 0.711 & 2.67& 0.383 & 0.865 &0.201\\
		3DPPE~\cite{shu20233dppe} & ICCV'23 & R101& 45.8 & 39.1 & & & & &\\
            BEVFormerV2~\cite{yang2023bevformerv2} & CVPR'23 & V299& 46.7 & 39.6 & 0.709 & 0.274 & 0.368 & 0.768 &0.196\\
		CMT-C~\cite{yan2023cross} & ICCV'23 & V299& 46.0 & 40.6 & -& -& -& -&-\\
  \midrule
		UVTR-C+UniPAD$^\dagger$\cite{yang2024unipad} & CVPR'24 & ConvNeXt-S & 46.4 & 41.0 & 0.671 & 0.277 & 0.382 & 0.867 &0.211\\
  		\textbf{UVTR-C+GP} & - & ConvNeXt-S& \textbf{47.2} & \textbf{41.7} & 0.676 &0.278 &0.394 &0.815 & 0.200\\
    
            \midrule
            {StreamPETR~\cite{wang2023exploring}} & ICCV'2023 &R50   & 47.9   & 38.0 &0.686&0.280&0.622&0.303&0.217\\
             \textbf{StreamPETR+GP}  &-& R50 &\textbf{ 48.8} & \textbf{38.6} &0.671& 0.273&0.593& 0.307&0.206\\
		\bottomrule
	\end{tabular}
 }  
        \vspace{-0.7em}
 	\caption{
		 \textbf{3D Object Detection.}
		We compare with previous SOTA methods {\em without} test-time augmentation on the nuScenes {\em val} set.\\
		$\dagger$: denotes our reproduced results based on  MMDetection3D~\cite{mmdet3d2020}.
		C denotes the experiment under camera-only.
	}
    \vspace{-0.5em}
	\label{tab:nuscene_val}
\end{table*}


\begin{table*}[ht]
\begin{center}
\resizebox{\textwidth}{!}{%
\begin{tabular}{lcclc|cccc}
\hline
Method & Modality & Backbone &  Pretrain&Epochs & mAP  & AP$_{ped}$ & AP$_{divider}$ & AP$_{boundary}$ \\
\toprule
HDMapNet\cite{li2022hdmapnet} & C & Effi-B0\cite{koonce2021efficientnet} &  ImageNet&30 & 23.0  & 14.4 & 21.7 & 33.0 \\
HDMapNet\cite{li2022hdmapnet} & L & PointPillars\cite{lang2019pointpillars} &  ImageNet&30 & 24.1 & 10.4 & 24.1 & 37.9 \\
HDMapNet\cite{li2022hdmapnet} & C \& L & Effi-B0 \& PointPillars &  ImageNet&30 & 31.0  & 16.3 & 29.6 & 46.7 \\
\midrule
VectorMapNet\cite{liu2023vectormapnet} & C & R50&  ImageNet&110 & 40.9  & 36.1 & 47.3 & 39.3 \\
VectorMapNet\cite{liu2023vectormapnet} & L & PointPillars &  ImageNet&110 & 34.0  & 25.7 & 37.6 & 38.6 \\
VectorMapNet\cite{liu2023vectormapnet} & C \& L & R50 \& PointPillars &  ImageNet&110 & 45.2  & 37.6 & 50.5 & 47.5 \\
\midrule
MapTR-tiny$^\dagger$\cite{liao2022maptr} & C &R50 &  ImageNet&24 & 49.9& 52.0 & 45.3 & 52.4  \\
\textbf{MapTR-tiny$^\dagger$$+$GP} & C & R50 & Ours&{24} & \textbf{{51.8}}& \textbf{54.9 }& \textbf{45.8} &\textbf{54.9 }\\
\bottomrule
\end{tabular}
}
\end{center}
\vspace{-1.5em}
\caption{\textbf{HD-Map construction}. Comparisons with state-of-the-art methods on nuScenes \textit{val} set.  ``C'' and ``L'' respectively denotes camera and LiDAR.
The APs of HDMapNet and VectorMapNet are taken from the paper of MapTR\cite{liao2022maptr}. GP presents GaussianPretrain.}
\label{tab:HD-map_val}
\vspace{-0.5em}
\end{table*}

\definecolor{nbarrier}{RGB}{255, 120, 50}
\definecolor{nbicycle}{RGB}{255, 192, 203}
\definecolor{nbus}{RGB}{255, 255, 0}
\definecolor{ncar}{RGB}{0, 150, 245}
\definecolor{nconstruct}{RGB}{0, 255, 255}
\definecolor{nmotor}{RGB}{200, 180, 0}
\definecolor{npedestrian}{RGB}{255, 0, 255}
\definecolor{ntraffic}{RGB}{255, 240, 150}
\definecolor{ntrailer}{RGB}{135, 60, 0}
\definecolor{ntruck}{RGB}{255, 0, 0}
\definecolor{ndriveable}{RGB}{213, 213, 213}
\definecolor{nother}{RGB}{139, 137, 137}
\definecolor{nsidewalk}{RGB}{75, 0, 75}
\definecolor{nterrain}{RGB}{150, 240, 80}
\definecolor{nmanmade}{RGB}{160, 32, 240}
\definecolor{nvegetation}{RGB}{0, 175, 0}
\definecolor{nothers}{RGB}{0, 0, 0}

\begin{table*}[ht]
	\footnotesize
 	\setlength{\tabcolsep}{0.0025\linewidth}
	
	\newcommand{\classfreq}[1]{{~\tiny(\nuscenesfreq{#1}\%)}}  %
    \begin{center}

	\begin{tabular}{l|c|c|c| c c c c c c c c c c c c c c c c c}
		\toprule
		Method
		& \makecell{Input \\ Modality}& \makecell{Image \\ Backbone} & mIoU
        & \rotatebox{90}{\textcolor{ncar}{$\blacksquare$} car}& \rotatebox{90}{\textcolor{nbus}{$\blacksquare$} bus}& \rotatebox{90}{\textcolor{nbicycle}{$\blacksquare$} bicycle}
		
		& \rotatebox{90}{\textcolor{nbarrier}{$\blacksquare$} barrier}& \rotatebox{90}{\textcolor{nvegetation}{$\blacksquare$} vegetation}& \rotatebox{90}{\textcolor{nconstruct}{$\blacksquare$} const. veh.}

		& \rotatebox{90}{\textcolor{nmotor}{$\blacksquare$} motorcycle}

		& \rotatebox{90}{\textcolor{npedestrian}{$\blacksquare$} pedestrian}

		& \rotatebox{90}{\textcolor{ntraffic}{$\blacksquare$} traffic cone}

		& \rotatebox{90}{\textcolor{ntrailer}{$\blacksquare$} trailer}

		& \rotatebox{90}{\textcolor{ntruck}{$\blacksquare$} truck}

		& \rotatebox{90}{\textcolor{ndriveable}{$\blacksquare$} drive. suf.}

		& \rotatebox{90}{\textcolor{nother}{$\blacksquare$} other flat}

		& \rotatebox{90}{\textcolor{nsidewalk}{$\blacksquare$} sidewalk}

		& \rotatebox{90}{\textcolor{nterrain}{$\blacksquare$} terrain}

		& \rotatebox{90}{\textcolor{nmanmade}{$\blacksquare$} manmade}

		& \rotatebox{90}{\textcolor{nothers}{$\blacksquare$} others}\\
		\midrule
        MonoScene~\cite{cao2022monoscene} & Camera & R101-DCN & 6.06 & 9.38 & 4.93 & 4.26 & 7.23 & 7.65& 5.67 & 3.98 & 3.01 & 5.90 & 4.45 & 7.17 & 14.91 & 6.32 & 7.92 & 7.43 & 1.01 & 1.75\\
        BEVDet~\cite{huang2021bevdet} & Camera & R101-DCN & 11.73 & 12.97 & 4.18 & 0.0 & 15.29 & 15.26& 1.35 & 0.0 & 0.43 & 0.13 & 6.59 & 6.66 & 52.72 & 19.04 & 26.45 & 21.78 & 14.51 & 2.09\\
  	TPVFormer~\cite{li2022bevformer} & Camera & R101-DCN & 27.83 & 45.90&40.78&13.67& 38.90 & 16.78 & 17.23& 19.99& 18.85& 14.30& 26.69& 34.17& 55.65& 35.47& 37.55& 30.70& 19.40& 7.22\\
        CTF-Occ~\cite{tian2023occ3d} & Camera & R101-DCN & 28.53 & 42.24 & 38.29 & 20.56 & 39.33 & 18.0& 16.93 & 24.52 & 22.72 & 21.05 & 22.98 & 31.11 & 53.33 & 33.84 & 37.98 & 33.23 & 20.79 & 8.09\\
        \midrule
        BEVFormer~\cite{li2022bevformer} & Camera & R101-DCN & 23.67 & 41.09 & 34.41 & 9.98 & 38.79 & 14.46 & 13.24 & 16.51 & 18.50 & 17.83 & 18.66 & 27.70 & 48.95 & 27.73 & 29.08 & 25.38 & 15.41 & 5.03\\
        BEVFormer+\textbf{GP}& Camera & R101-DCN & 24.21 & 39.18 & 36.55 & 6.95 & 34.88 & 18.16  & 11.88 & 16.62 & 16.93 & 17.1 & 13.83 & 27.03 & 54.09 & 32.36 & 33.02 & 27.05 & 20.39 & 5.53\\
        \midrule
        PanoOCC\cite{wang2024panoocc} &  Camera & R101-DCN & 41.60 & 54.78 & 45.46 & 28.92 & 49.82 & 40.10 & 25.20 & 32.93 & 28.86 & 30.71 & 33.87 & 41.32 & 83.18 & 45.00 & 53.80 & 56.10 & 45.11 & 11.99\\
        PanoOCC+\textbf{GP} &  Camera & R101-DCN & \textbf{42.42} & 55.21 & 49.88 & 28.81 & 49.30 & 42.54  & 22.27 & 31.30 & 29.42 & 30.37 & 34.29 & 42.05 & 84.06 & 47.76 & 55.90 & 58.13 & 48.20 & 11.58\\

		\bottomrule
	\end{tabular}
    \end{center}
    \vspace{-2em}
    \caption{\textbf{3D Occupancy prediction performance on the Occ3D-nuScenes dataset.} * means the performance is achieved by using the camera mask during training. GP is an abbreviation for GaussianPretrain.}
 \vspace{-10pt}
	\label{tab:camera_occ}
\end{table*}

\subsection{\textbf{Supervise by Reconstruction Signals}}
\label{sec:losses}


To facilitate a better reconstruction of the masked region under the MAE strategy, we supervise the learning process with different reconstruction signals derived from the Gaussian representation. Specifically, the \textit{RGB}, \textit{Depth}, and \textit{Occupancy} signals are decoded based on the predicted Gaussian anchor parameters within the valid mask patches.

\paragraph{\textbf{RGB Reconstruction.}}
Since we do not need to reconstruct images of arbitrary perspectives, we directly predict fixed viewpoint  \textit{RGB} instead of using \textit{Spherical Harmonics {(SH)}} coefficients.
After predicting the parameters of the Gaussian anchor, we decode the color information with the  \cref{reder_rgb} to render the RGB values map of the image, denoted as $\hat{C}$, for each target reconstruct pixel. Specifically, the $c_i$ value in the equation is replaced by the predicted RGB.
\vspace{-1em}
\paragraph{\textbf{Depth Reconstruction.}}
Inspired by the depth implementation in NeRF-style volume rendering, we integrate the depth of each splat in a manner similar to RGB reconstruction. We approximate the each pixel z-depth from the 3D-GS parameters. The process is as follows:
\begin{equation}
\hat{D} =\sum_{i=1}^n d_{i}\alpha_{i} \prod^{i-1}_{j=1} (1-\alpha_{j} )
\end{equation}
where n is the number of Gaussian anchors, $d_{i}$ is the $i^{th}$ Gaussian anchor $z$-depth coordinate in view space, enabling efficient depth rendering with minimal computational overhead. $\hat{D}$ is the depth map of the image.  
%
\vspace{-1em}
\paragraph{\textbf{Occupancy Reconstruction.}}
The opacity attribute of 3D-GS point is inherent to vision perception, particularly for occupancy prediction tasks.
Unlike GaussianFormer\cite{huang2024gaussianformer}, which uses opacity for semantic logits, we directly interpret opacity as an indicator of occupancy. 
%
A Gaussian anchor with full opacity signifies the presence of an occupied location at $x$.  Formally, for each target voxel, we take the maximum opacity value among Gaussian anchors within the voxel to represent the occupancy probability, denoted by $\hat{O}$. This direct mapping of opacity to occupancy provides a natural and effective way to leverage 3D Gaussian Splatting for occupancy prediction.
\begin{equation}
\hat{O} = \max_{j=1}^{k}{(\mathcal{M}^j_\alpha(x)) \mid x \in V_t}
\end{equation}
where $k$ is number of Gaussian anchors in a target voxel $V_t$.
\vspace{-2em}
\paragraph{\textbf{Loss Function.}}

In summary, the overall pre-training loss function consists of color loss, depth loss and occupancy loss:
\begin{equation}
\begin{aligned}
\resizebox{\linewidth}{!}{$
\begin{gathered}L=\frac{\lambda_{RGB} }{N_t^p} \sum^{N_t^p}_{i=1} \left| C_{i}-\hat{C}_{i} \right|  +\frac{\lambda_{Depth} }{N_t^p}  \sum^{N_t^p}_{i=1} \left| D_{i}-\hat{D}_{i} \right|  \\ +\frac{\lambda_{Occupacy} }{N_t^v} \sum^{N_t^v}_{i=1} \left| O_{i}-\hat{O}_{i} \right|  \end{gathered} 
$}.
\end{aligned}
\end{equation}
where $C_{i}$, ${D}_{i}$ are the ground-truth color, depth for each ray. $O_{i}$ denotes ground-truth of occupancy which is considered occupied if it contains at least one lidar point. $N_t^p$ and $N_t^v$ are the counts of the target pixels of $P_t$ and target voxels of $V_t$, respectively.

\section{Experiments}
\label{sec:experiments}
\subsection{\textbf{Datasets and Evaluation Metrics}} 

\noindent \textbf{nuScenes dataset}~\cite{caesar2020nuscenes} contains 700/150/150 scenes for training, validation, and testing, respectively. Each sequence is captured at 20Hz frequency with 20 seconds duration.
Each sample contains RGB images from 6 cameras with 360$^{\circ}$ horizontal FOV and point cloud data from 32 beam LiDAR sensor, and five radars.  For HD-map task, perception ranges are $[-15.0m, 15.0m]$ for the $X$-axis and $[-30.0m, 30.0m]$ for the $Y$-axis. We calculate the $\rm{AP}_{\tau}$ under Chamfer distance with several thresholds ($\tau \in T, T=\{0.5, 1.0, 1.5\}$).

\noindent \textbf{Occ3D-nuScenes}~\cite{tian2023occ3d} occupancy scope is defined as $-40m$ to $40m$ for X and Y-axis, and $-1m$ to $5.4m$ for the Z-axis. The voxel size is $0.4m \times 0.4m \times 0.4m$ for the occupancy label. Occ3D-nuScenes benchmark calculates the mean Intersection over Union (\textbf{mIoU}) for 17 semantic categories (including `others'). Besides, it also provides visibility masks for LiDAR and camera modality.

\subsection{\textbf{Implementation Details}}
Our code implementation is based on the MMDetection3D codebase, and all models were trained on \textbf{8} NVIDIA A100 GPUs. Unless otherwise specified, the input image resolution is set to 1600x900 by default. The scale factors for $\lambda_{RGB}$ and $\lambda_{Occ}$ are maintained at 10, while $\lambda_{Depth}$ is set to 1. During the pre-training phase, we apply a mask to the input images, with a mask size of 32 and a ratio of 0.3. LiDAR specific depth range for generating the valid mask is [0, 50]. The model is pre-trained for 12 epochs using the AdamW optimizer, with an initial learning rate of 2e-4 and a weight decay of 0.01, without using CBGS [78] or any data augmentation strategies. In the ablation studies, 
unless explicitly stated, fine-tuning is conducted for 12 epochs on 50\% of the image data.

\subsection{\textbf{Main Results on Different Vision Task}} 

\paragraph{3D Object Detection.}
We compare our GaussianPretrain with previous SOTA methods as shown in \cref{tab:nuscene_val}. Taking the pre-training framework, UniPAD, as baseline which was achieved on UVTR-C/StreamPETR. Our method outperforms UniPAD-C over {0.8} and {0.7} points on NDS and mAP, respectively. The improvement further gained {0.9} NDS compared to StreamPETR, achieved {48.8} and {38.6} on NDS and mAP, reaching the level of existing state-of-the-art methods without any test time augmentation.
\vspace{-1.0em}
\paragraph{HD Map Construction.}
As shown in \cref{tab:HD-map_val}, we evaluate the performance of our pre-training model on the nuScenes dataset for the HD map construction task. This task requires the model to understand road topology and traffic rules, necessitating a detailed understanding of the scene's texture information. We utilize MapTR~\cite{liao2022maptr} to assess the ability of GaussianPretrain to capture this information. Benefiting from our effective pre-training of Gaussian representation, MapTR achieves a \textbf{1.9\%} improvement in mAP.

\noindent{\textbf{3D Occupancy Prediction.}}
The opacity attribute of Gaussian anchor is inherently suited for occupancy prediction tasks. In \cref{tab:camera_occ}, we conduct the experiments of 3D occupancy prediction on the Occ3D-nuScenes. 
The performance of the SOTA methods in the table 
is reported in the work of Occ3d~\cite{tian2023occ3d}. 
We implement our framework on  BEVFormer~\cite{li2022bevformer} and PanoOCC\cite{wang2024panoocc}, achieving an improvement of {0.6}\% mIoU over BEVFormer and a further improvement of {0.8}\% mIoU over the SOTA method, PanoOCC. This also highlights the effectiveness of our pre-training diagram.

\begin{table}[h]
    \centering    
    
\resizebox{0.95\columnwidth}{!}
{
\begin{tabular}{l|c|l|cc}
\hline
\textbf{Base Model} &  Backbone&\textbf{Pretrain}  & \textbf{NDS$\uparrow$} & \textbf{mAP$\uparrow$} \\ \hline
&  ConvNeXt-S&DD3D   & 26.9    & 25.1  \\
&  ConvNeXt-S&SparK    & 29.1   & 28.7  \\
 &  ConvNeXt-S&FCOS3D  & 31.7 & 29.0       \\
UVTR-C&  ConvNeXt-S&UniPAD  & 31.0  & 31.1  \\ \cline{2-5}
&ConvNeXt-S&ImageNet   & 25.2   & 23.0     \\
 &  ConvNeXt-S&Ours & \textbf{32.3} & \textbf{32.0}        \\ 
\bottomrule
\end{tabular}
}
\vspace{-0.7em}
\caption{Performance of different pre-training methods.}
    \label{tab:different-pretrain}
\end{table}

\subsection{\textbf{Main results in Pre-training Methods}} 
\noindent \textbf{Comparisons with Pre-training Methods.} We take UVTR-C pre-trained on ImageNet as baseline and validate our GaussianPretrain by comparing it with previous pre-training methods in \cref{tab:different-pretrain}. 1) DD3D: utilizes depth estimation for pre-training. 2) SparK: incorporates the MAE to pre-training method. 3) FCOS3D: employs 3D labels for supervision during the pre-training phase. 4) UniPAD: NeRF-based rendering pre-training paradigm. Our GaussianPretrain, which integrates 3D-GS into vision pre-training, largely improved by 7.1\% in NDS and 9.0\% in mAP over the baseline.  It outperforms all other methods,  achieving 32.0 NDS and 32.3 mAP, respectively.

\begin{table}[!t]
    \centering
    \resizebox{\columnwidth}{!}
    {
        \begin{tabular}{l|c|c|cc}
        \toprule
         \textbf{Models} &\textbf{Backbone} & 
         \textbf{Pretrain} & 
         \textbf{NDS$\uparrow$} & 
         \textbf{mAP$\uparrow$} \\
        \bottomrule
            \multirow{2}{*}
            {UVTR-C~\cite{li2022unifying}}  
            &ConvNeXt-S & ImageNet  & 29.20 & 26.54 \\
            &ConvNeXt-S & GaussianPretrain  & 36.90 & 35.78 \\
             \hline
            \multirow{2}{*}{BEVFormer~\cite{li2022bevformer}}   &R101 & ImageNet  & 44.30 & 32.70 \\
            &R101 & GaussianPretrain  & 45.76 & 34.73 \\
             \hline
             \multirow{2}{*}{StreamPETR~\cite{wang2023exploring}} &R50  & ImageNet  & 47.94   & 38.04 \\
                                         &R50 & GaussianPretrain  & 48.79 & 38.59 \\
        \bottomrule
        \end{tabular}
    }
    \vspace{-0.7em}
\caption{Effectiveness on different detection models. }
 \vspace{-1.0em}
    \label{tab:Comparsion-3DDet}
\end{table}
\noindent \textbf{Implement on Diverse Models.}
To demonstrate the effectiveness of GaussianPretrain, we implement our method on different models, including UVTR-C~\cite{li2022unifying}, StreamPETR~\cite{wang2023exploring} and BEVFormer~\cite{li2022bevformer}, under different backbones as shown in ~\cref{tab:Comparsion-3DDet}. Notably, GaussianPretrain enhances the performance across all models. Specifically, UVTR-C~\cite{li2022unifying} exhibits improvements of \textbf{7.7\%} in NDS and \textbf{9.24\%} in mAP. BEVFormer~\cite{li2022bevformer} shows gains of \textbf{1.46\%} in NDS and \textbf{2.03\%} in mAP. Even the highly competitive temporal 3D detection model, StreamPETR~\cite{wang2023exploring}, benefits from our pre-training, achieving 0.85\% in NDS.
Specifically, BEVFormer and StreamPERT are fine-tuned on the full image dataset for 24 epochs, while UVTR is trained for 12 epochs.

\begin{table*}[th!]
 \centering
\fontsize{6pt}{8pt}\selectfont 
 \resizebox{0.85\textwidth}{!}
 {
\begin{tabular}{llcccccccccc}
  \toprule
  \multirow{2}{*}{Method} & \multirow{2}{*}{Modality} & \multicolumn{3}{c}{Distance: NDS(\%)} & & \multicolumn{2}{c}{Weather: NDS(\%)} & & \multicolumn{2}{c}{Lighting: NDS(\%)} \\ \cline{3-5} \cline{7-8} \cline{10-11}
  & & \textless20{\em m} & 20-30{\em m} & \textgreater30{\em m} & & {\em Sunny} & {\em Rainy} & & {\em Day} & {\em Night} \\ 
  \midrule
UVTR-C & Camera & 33.1 & 20.8 & 13.8 & & 24.2 & 21.2 & & 25.1 & 20.8 \\
UniPAD-C & Camera & 39.0 & 27.3 & 15.8 & & 29.8 & 26.1 & & 31.0 & 25.4 \\
  \midrule
  GaussianPretrain & Camera & 39.1 & 28.9 & 16.9 & & 30.8 & 27.1 & & 32.1 & 26.5 \\
  \bottomrule
 \end{tabular}
 }
 \vspace{-0.8em}
\caption{Comparisons of different distances, weather, and lighting conditions on nuScenes {\em val} set.
 }
 \label{tab:condition} 
\end{table*}
\subsection{\textbf{Ablation Studies.}}

\begin{table}[t]
\vspace{-0.5cm}

    \centering
    \resizebox{0.9\columnwidth}{!}{
    \small 
        \begin{tabular}{c|c|c|cc|c}
        \toprule
        \textbf{RGB-Loss} & \textbf{Depth-Loss} & \textbf{OCC-Loss} &  \textbf{NDS$\uparrow$} & \textbf{mAP$\uparrow$}  &\textbf{{mIoU}}\\
        \bottomrule
          \ding{55}   & \ding{55}    & \ding{55}  & 25.23  & 23.00& 15.1\\
$\checkmark$  &       &               & 26.84 & 25.73  & 16.3\\
$\checkmark$  & $\checkmark $   &               & 29.20 & 26.54  & 17.2\\
 \rowcolor[gray]{.92}
$\checkmark$  & $\checkmark $   & $\checkmark $ & \textbf{32.28} & \textbf{31.99}  & \textbf{19.3}\\  \bottomrule
\end{tabular}
 }
 \vspace{-0.7em}
\caption{Ablation study on different supervised losses.}
\label{tab:diff_loss}
\end{table}

\paragraph{Effect of GaussianPretrain's losses.}
To validate the effectiveness of each reconstructed signal, we conducted experiments on UVTR and BEVFomer for 3D detection and occupancy tasks, respectively. The RGB loss guides the model in learning texture information of the scene from the reconstructed image, while the depth loss encourages the model to learn the geometric information on a 2D plane, although this alone is insufficient for capturing complete 3D geometry. In contrast, the occupancy loss supervises the model in learning comprehensive geometric information within 3D space. As shown in \cref{tab:diff_loss}, we experiment on 1/2 subset for 12 epochs, each component contributes positively, with the best results achieved when all are used together. Notably, thanks to the attribute of the opacity of Gaussian representation which inherently benefits occupancy prediction, yielding a significant improvement of \textbf{ 2.1\%} in mIoU. These results underscore the effectiveness of Gaussian representation to the pre-training framework.

\begin{table}[!t]
\vspace{-0.2cm}
    \centering
    \resizebox{0.9\columnwidth}{!}{
    \small 
        \begin{tabular}{c|c|c|c|c}
        \toprule
        \textbf{Numbers}& \textbf{latency(ms)}& \textbf{ Memory(MB)}&  \textbf{NDS$\uparrow$}&  \textbf{mAP$\uparrow$}\\
        \bottomrule
256x100&         16&      502&           31.17   &   30.94\\
512x100&        17&       608&           31.70   &   31.55\\
 \rowcolor[gray]{.92}
1024x100&       19&       788&           32.28   &   31.99\\
2048x100&       25&       1170&           \textbf{32.42}   &   \textbf{32.08}\\
        \bottomrule
        \end{tabular}
    }
    \vspace{-0.7em}
    \caption{Ablation study on the number of Gaussian anchors.}
    \label{tab:gaus_parms}
\end{table}
\vspace{-1em}
\paragraph{Ablation on Gaussian Anchor Numbers.} We conducted an ablation study to examine the effect of varying the number of Gaussian anchors on performance metrics as shown in \cref{tab:gaus_parms}. The most significant gains are observed up to 1024 rays, beyond which improvements are smaller relative to the additional resource demands. In this study, we define the number of Gaussian anchors per ray as \textit{100}.

\begin{table}[!t]
\vspace{-0.2cm}

    \centering
    \resizebox{0.9\columnwidth}{!}{
    \small 
        \begin{tabular}{c|c|l|c|c}
        \toprule
        \textbf{Method}& Decoder &Param& Memory&  Latency\\ \bottomrule  
UniPAD-C&  NeRF&   0.46MB &    1125MB& 32ms\\
GaussianPretrain&  3D-GS& 0.45MB   & 788MB &19ms\\
        \bottomrule
        \end{tabular}
    }
    \vspace{-0.8em}
    \caption{Comparison the consumption with NeRF-based method.}
    \vspace{-0.8em}
    \label{tab:consumption}
\end{table}
\vspace{-1em}
\paragraph{Efficiency \& consumption.}
NeRF-based methods often suffer from slow convergence and high GPU memory consumption. In contrast, our 3D-GS-based approach offers comparable rendering quality with significantly faster convergence and superior efficiency for free-view rendering. In the \cref{tab:consumption}, we compare the efficiency and memory consumption of the decoder module between the NeRF-base UniPAD and ours. Notably, GaussianPretrain obviously reduces memory usage by about \textbf{30\%}, and decreases latency by approximately \textbf{40.6\%}, while maintaining a similar parameter size. This highlights significant gains in resource efficiency with the GaussianPretrain approach.
\begin{figure}[t]
\vspace{-0.55cm}

  \centering
  \includegraphics[width=0.5\textwidth]{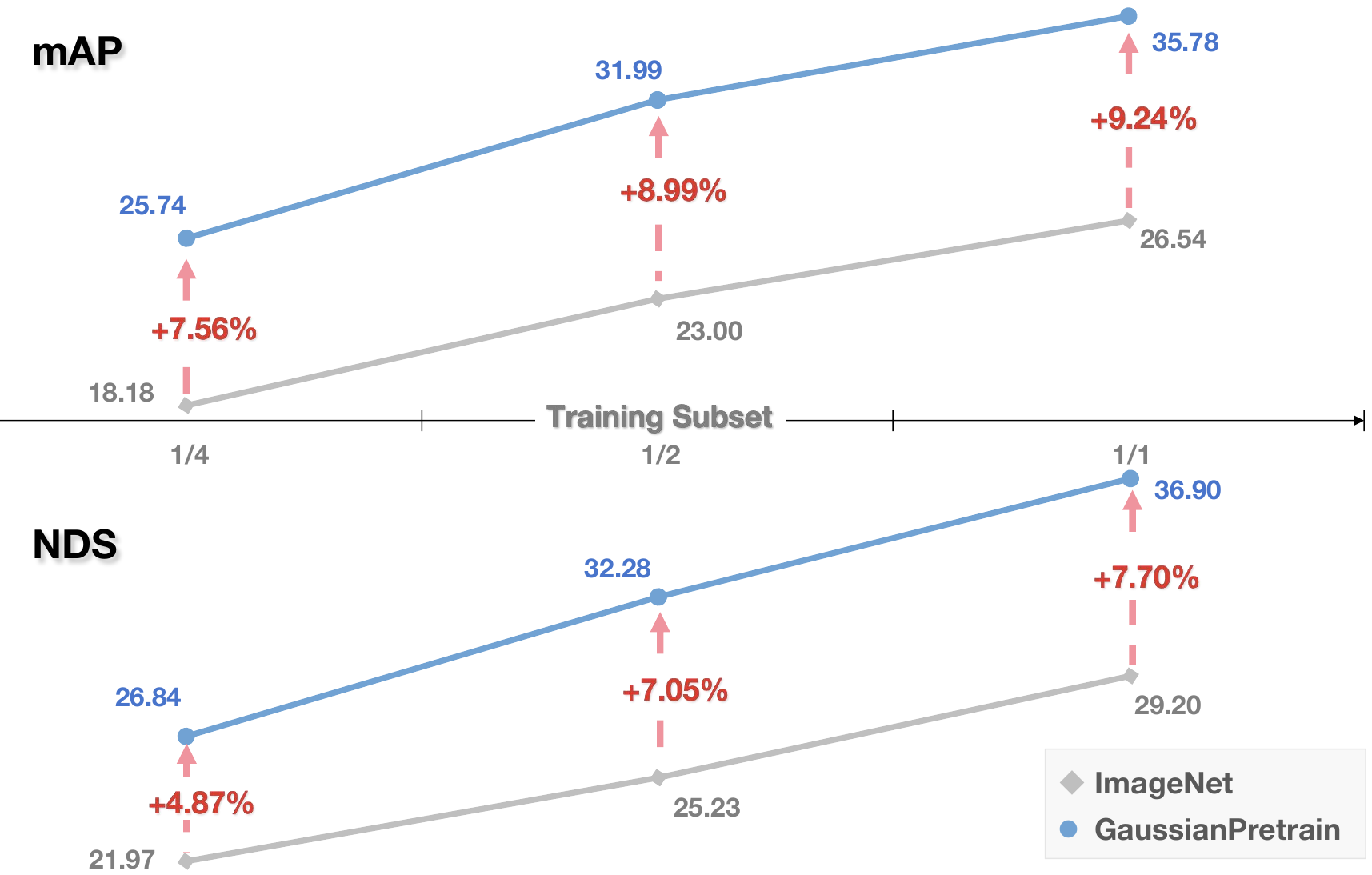}
    \caption{
  Effect of GaussianPretrain on Fine-tuning. By reducing annotations from the full training set to a 1/4 subset).
}
  \label{fig:training-stage}
\end{figure}

\vspace{-1.1em}
\paragraph{\textbf{Effect of Supervised Pre-training.}} 
We demonstrate the effectiveness of GaussianPretrain in reducing the reliance on annotations by fine-tuning UVTR~\cite{li2022unifying}, ranging from the full dataset to a 1/4 subset. As depicted in ~\cref{fig:training-stage}, our approach surpasses the baseline under full supervision by 5.5\% mAP, with only half of the supervised samples, i.e., 32.0\% mAP vs. 26.5\% mAP. This result indicates that GaussianPretrain can effectively leverage unlabeled data to compensate for reduced supervision, leading to improved performance even with fewer annotations.

\vspace{-1.0em}
\paragraph{Different Conditions.} 
We report the performance with different distances,
weather conditions, and light situations in the \cref{tab:condition}, which benefits from our effective pre-training on UVTR-C, GaussianPretrain achieves superior robustness and overall the best performance.



\section{Conclusion and Limitations.}
\vspace{-0.8em}
In this work, we introduce 3D Gaussian Splatting technology into vision pre-training task for the first time. Our GaussianPretrain demonstrates remarkable effectiveness and robustness, achieving significant improvements across various 3D perception tasks, 
 including 3D object detection, HD map reconstruction, and occupancy prediction, 
with efficiency and lower memory consumption.

\vspace{0.8em}
\noindent \textbf{Limitation.} There still exist certain limitations in the current framework. Notably, it does not explicitly incorporate temporal or multi-modality information, both of which are crucial for many autonomous driving applications. In future work, we plan to extend GaussianPretrain to leverage this information and further enhance its performance.
{
    \small
    \bibliographystyle{ieeenat_fullname}
    \bibliography{main}
}


\end{document}